\documentclass[letterpaper, 10 pt, conference]{IEEEconf}

\IEEEoverridecommandlockouts                              

\usepackage{amsmath} 
\usepackage{amssymb}  
\usepackage{graphicx}
\usepackage{amsfonts}
\usepackage{bm}
\usepackage{nccmath}
\usepackage{mathtools}
\usepackage{hyperref}
\def\BibTeX{{\rm B\kern-.05em{\sc i\kern-.025em b}\kern-.08em
    T\kern-.1667em\lower.7ex\hbox{E}\kern-.125emX}}
    \hypersetup{
    colorlinks=true,
    linkcolor=blue,
    filecolor=magenta,      
    urlcolor=cyan,
}

\newcommand{\R}{\mathbb{R}}

\newcommand{\bmat}[1]{\begin{bmatrix} #1 \end{bmatrix}}

\newcommand{\pmat}[1]{\begin{pmatrix} #1 \end{pmatrix}}

\newcommand{\p}{\partial}
\newcommand{\pdif}[2]{\frac{\p #1}{\p #2}}

\newcommand{\mpdif}[2]{\mfrac{\p #1}{\p #2}}

\mathchardef\mhyphen="2D

\newcommand{\convhull}{\operatorname{convhull}}
\newcommand{\diag}{\operatorname{diag}}

\newcommand{\bM}{\mathbf}

\title{\LARGE \bf
The dynamic effect of mechanical losses of transmissions on the equation of motion of legged robots
}

\author{Youngwoo Sim$^1$ and Joao Ramos$^1$
\thanks{$^{1}$Authors are with the Department of Mechanical Science and Engineering at the University of Illinois at Urbana-Champaign, Urbana, IL 61801, USA. Corresponding author: {\tt\small sim17@illinois.edu}}
\thanks{This work is supported by the National Science Foundation via grant IIS-2024775}
}

\begin{document}
\maketitle

\thispagestyle{empty}
\pagestyle{empty}

\begin{abstract}

Industrial manipulators do not collapse under their own weight when powered off due to the friction in their joints. Although these mechanism are effective for stiff position control of pick-and-place, they are inappropriate for legged robots that must rapidly regulate compliant interactions with the environment. However, no metric exists to quantify the robot's performance degradation due to mechanical losses in the actuators and transmissions. 
This paper provides a fundamental formulation that uses the mechanical efficiency of transmissions to quantify the effect of power losses in the mechanical transmissions on the dynamics of a whole robotic system. 
We quantitatively demonstrate the intuitive fact that the apparent inertia of the robots increase in the presence of joint friction. We also show that robots that employ high gear ratio and low efficiency transmissions can statically sustain more substantial external loads. We expect that the framework presented here will provide the fundamental tools for designing the next generation of legged robots that can effectively interact with the world.

\end{abstract}

\section{Introduction}
The mechanical losses in the transmissions of individual joints govern the system-level dynamics of robots. For instance, conventional industrial manipulators behave like statues when they are powered off; in other words, they are \textit{non-backdrivable}. The characteristics that determine this behavior are the low mechanical efficiency and high friction in the gearboxes. Although these mechanical transmissions have been successfully utilized in industrial manipulators for stiff position control, they are not appropriate for compliant force control due to their large mechanical impedance \cite{ImpactMitigation}. Therefore, to enable legged robots to control their contact forces with the environment, one must analyze how the power efficiency of the mechanical transmissions utilized in robots governs the overall dynamics. Towards this goal, this paper introduces a framework for studying how the energetic losses at joint-level propagate to the dynamic behavior of the robot at system-level.

\begin{figure}[t]
  \centering
  \includegraphics[width =1\linewidth]{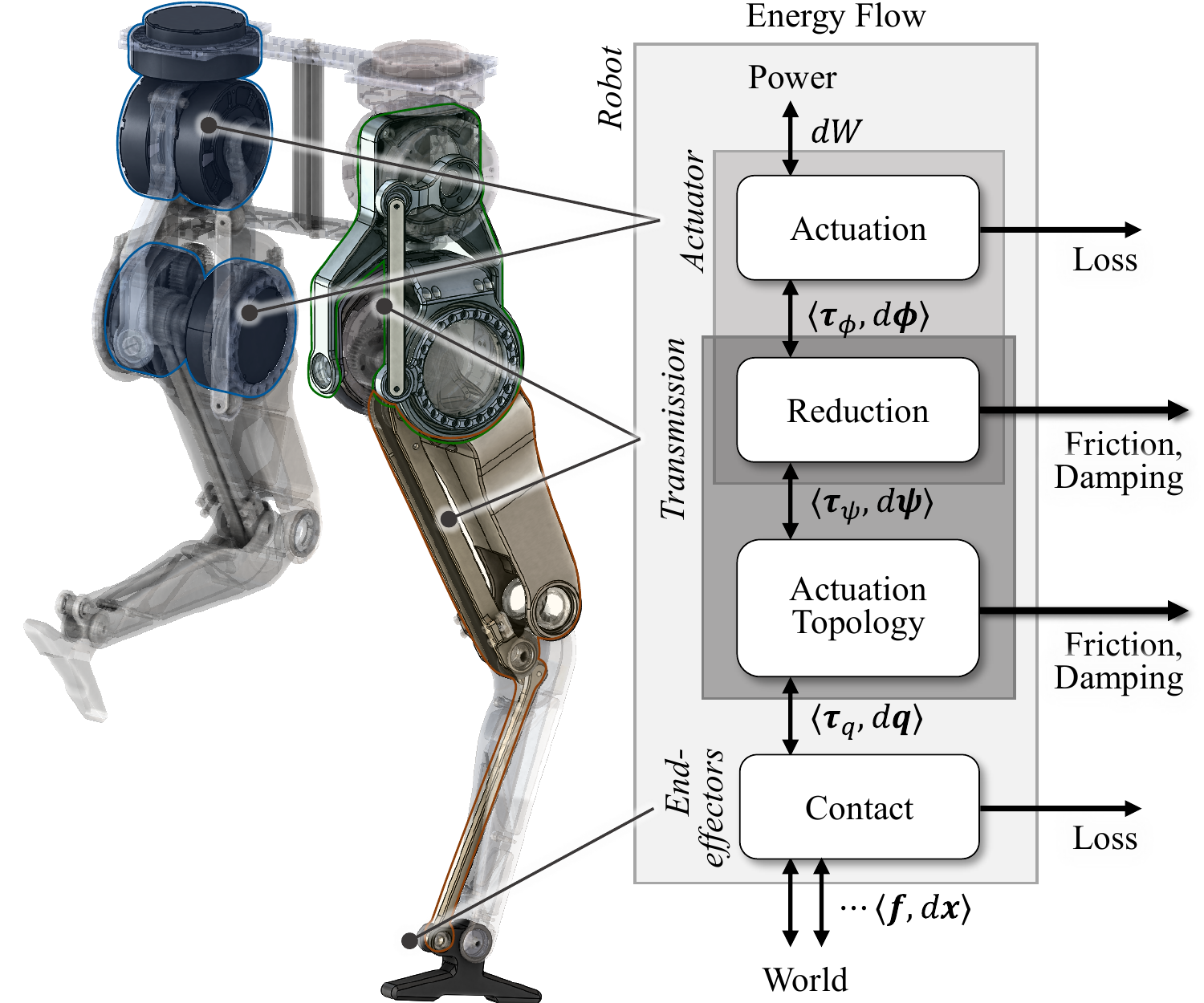}
  \caption{Energy flow diagram of a robotic system showing the dissipation of energy in  actuators and transmissions. The energy conversions are always accompanied by energy losses such as Joule heating or friction. }
  \label{Fig:TelloDiagram}
\end{figure}

No existing design metric can quantitatively describe the performance degradation of a robotic system due to the mechanical losses in actuators or transmissions. The absence of such metric hinders the selection of an optimal mechanical transmission for legged robots. For instance, no existing design guideline provides a clear choice between a compact and low efficiency strain wave gearbox and a bulkier, but higher efficiency, planetary gearbox used in many legged robots \cite{MiniCheetah,Ramos_LilHERMES}. 
This unanswered question arose in the design of  the humanoid robot \textit{TELLO}, shown in Fig. \ref{Fig:TelloDiagram} and in the video \cite{TELLO_Vid_Leg}, and motivated this study. 

Related work in the literature analysed the fundamental behavior of mechanical transmission and their impact on actuator backdrivability. 
Giberti provides the optimal choice of an actuator-reducer pair of one-degree-of-freedom (DoF) system, considering joint level efficiency \cite{ChoiceMotorReducer}. 
Similarly, Wang demonstrates that gearboxes present directional efficiency, which means that the mechanical losses are different if actuators operates within positive or negative work regimes \cite{Wang_DirectionalEfficiency}. Wensing investigates how joint-level apparent inertia decreases backdrivability, and thus, degrades the impact mitigation capability of the whole system \cite{ImpactMitigation}. Singh studies how the mass distribution of a robot's leg influences the propagation of impact from the ground to the torso \cite{SinghShockPropagation}. Kim demonstrated a light-weighted robot arm of high backdrivability with a novel tension-amplification transmission \cite{LIMS}. However, the influence of the mechanical efficiency of transmissions on the whole system still remains unstudied. 

\begin{figure*}[t]
  \centering
  \includegraphics[width =1\linewidth]{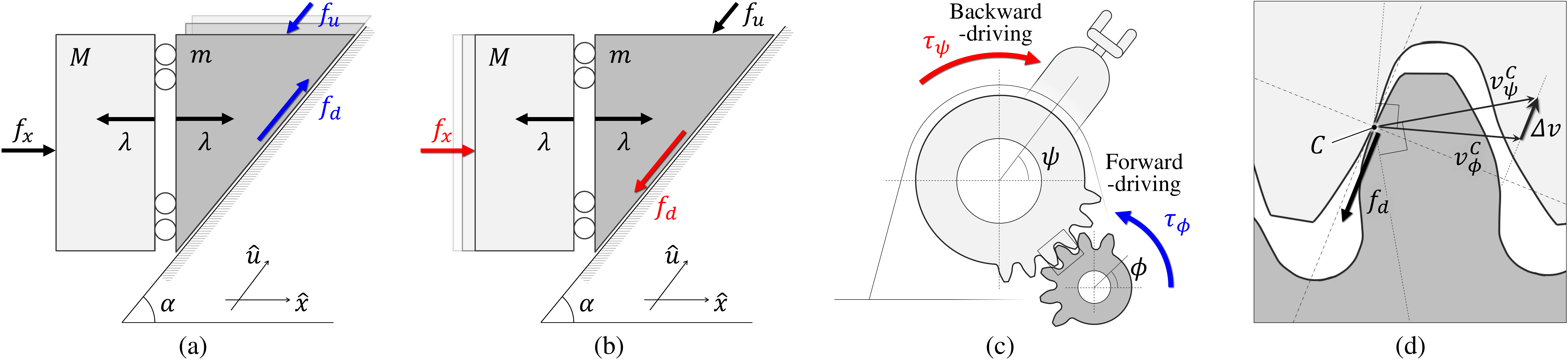}
  \caption{(a) A forward-driving scenario of a wedge-block model analogous to a geared transmission. (b) A backward-driving scenario (c)  The forward- and backward-driving scenarios of a rotor-manipulator model that represents a typical geared transmission (d) A dissipative force $f_d$ associated with the slip $\Delta v$ at the contact point $C$ of a typical geared transmission \cite{YadaGearReview}}
  \label{Fig:SlopeAnalogy}
\end{figure*}

The main contribution of this paper is to provide a fundamental formulation to quantify the effect of power losses in the mechanical transmissions on the dynamics of the whole robotic system. As a product of this formulation, we propose an augmented equation of motion that embeds the mechanical efficiency in the system's inertia matrix and the vectors of bias force, the sum of Coriolis and gravitational force, and generalized forces. The system-level impact of the individual efficiencies is demonstrated by computing the (task-space) generalized inertia ellipsoid \cite{AsadaInertiaEllipsoid} and force capability of the robot's end-effector. Two core results are obtained from this study. First, the perceived inertia of the robot at the end-effector increases as the efficiency decreases. Second, the capability of the robot to generate or resist force depends on the direction of the energy flow in the system.

This paper is organized as follows. First, we study a simple representative example to understand how transmission efficiency affects the dynamics of a one-DoF system in forward-driving (FWD) and backward-driving (BWD) scenarios. Next, the kinematic structure and constraints of rigid body system are highlighted to trace the energy loss inherent to the mechanical transmission. In Section IV, the generalized dynamics of a rigid-body system with mechanical power losses in kinematic structure  is obtained. The proposed equation of motion enables the re-derivation of a {conventional} design criteria, such as the generalized inertia tensor and the force capability. Finally, a case study of a 2-DoF leg {qualitatively} validates the proposed formulation.

\section{A Simple Model of dissipative dynamics}

This section investigates how the power efficiency of a mechanical transmission is embedded into the equation of motion using a simple \textit{wedge-block model} shown in Fig. \ref{Fig:SlopeAnalogy}(a) and (b). This model extracts the essential behavior of the complex sliding and pushing dynamics of the gear teeth meshing in the rotor-manipulator model in Fig. \ref{Fig:SlopeAnalogy}(c) and (d). The wedge-block model is designed to inherit mechanical characteristics of rotor-manipulator model; (i) frictional loss is dictated by the geometry of meshing between two bodies, and (ii) the dynamics is different depending on the direction of {energy flow.} 
This flow falls into two categories:
\begin{itemize}
    \item FWD (Fig. \ref{Fig:SlopeAnalogy}(a)) occurs when pushing the wedge $m$ with a force $f_u$ and, consequently, moving the block $M$. This is equivalent to commanding a motor torque $\tau_\theta$ to drive the link of a manipulator. 
    \item BWD (Fig. \ref{Fig:SlopeAnalogy}(b)) occurs when energy flows in the opposite direction by pushing the block $M$ with a force $f_x$ to drive the wedge $m$. This is equivalent to an external force applied to the manipulator's end-effector to backdrive the rotor through the mechanical transmission.
\end{itemize}  

The analogy between the wedge-block and rotor-manipulator models is described here. First, {the displacements of the block $x$ and the wedge $u$ are coupled by the slope with incline $\alpha$. In this case, $\tfrac{1}{\cos\alpha}$ is the mechanical advantage that increases with the slope angle. This is expressed by the constraint
\begin{align*}
    g(x, u) & = -x +({\cos\alpha}){u}  \quad (\alpha \in [0, \tfrac{\pi}{2}))
\end{align*}
Similarly, the angular displacements of the rotor $\phi$ and the manipulator joint $\psi$ are kinematically coupled by gear ratio ${N}$ which is the ratio of input velocity to output velocity.  
\begin{align*}
    g(\psi, \phi) &= -\psi + \tfrac{1}{N}\phi  \quad (N \ge 1 ).
\end{align*}}
Second, we study the dynamics of different energy flow scenarios, where the normal force $\lambda$ and sliding friction $f_d$ maintain their orientation and magnitude. However, the sign of the friction flips according to the movement of wedge, which contributes to the \textit{asymmetric} dynamic behavior in FWD and BWD cases. As a consequence, the example renders different mechanical efficiencies in the FWD and BWD scenarios, similarly to geared transmissions \cite{Wang_DirectionalEfficiency}. The power loss inherent to the gear meshing mechanics largely contributes to the asymmetry of the dynamics. This means that the power loss, mechanical efficiency, apparent inertia, and input power distribution are different if the motors are driving the manipulator or if an external force is back-driving the actuators. The rest of this section derives and discuss the dynamics of forward and back-driving cases of the wedge-block model in Fig. \ref{Fig:SlopeAnalogy}.


\subsection{{Meshing forces}}
Let the redundant coordinates be {$\bm s = [x \quad u]^\top$}. The dynamic equations of motion of the wedge-block model in Fig. \ref{Fig:SlopeAnalogy}(a) and (b) follows
\begin{gather*}
\bM H_s \ddot{ \bm s} - \bM A^\top \lambda = \bm f + \bm f_d, 
\end{gather*}
from standard Lagrangian formulation, with 
\begin{gather*}
\bM H_s \!=\! \bmat{M \!&\! 0 \\ 0 \!&\! m}\!\!,\,  \bM A^\top \!= \!\bmat{-1 \\ \cos\alpha}\!\!,\,  \bm f \!=\! \bmat{f_x \\ f_u}\!\!,\, \bm f_d \!=\! \bmat{0 \\ \pm \mu \sin\alpha \lambda}\!\!,
\end{gather*}
where $\bm f_d$ is the dissipative force and $\bM A =\pdif{g}{\bm s}$ is the constraint Jacobian that represents the mechanical advantage that distributes the constraint force. We introduce the concept of \textit{meshing forces} $\bm r$, which are the sum of dissipative forces $\bm f_d$ and the constraint forces $\bM A^\top \lambda$.
\begin{gather*}
\bm r \coloneqq \left(\bM A^\top \lambda + \bm{f}_d \right) = \bM H_s \ddot{\bm s} - \bm{f }. 
\end{gather*}
This rearrangement groups the contact forces that transmits power between bodies. {Writing the $x$ and $u$ components of the meshing force $\bm r$,} it becomes clear that the dry friction $\pm \mu\sin\alpha\lambda$ contributes to the asymmetricity of the dynamics:
\begin{align}
    \bmat{r_x \\ r_u} = \bmat{-1 \\ \cos \alpha (1\pm \mu \sin \alpha )}\lambda = \bmat{M\ddot x - f_x \\ m \ddot u -f_u}.
    \label{eqn:exDyn}
\end{align}
With \eqref{eqn:exDyn}, the mechanical efficiency $\eta$, the ratio of output power $(\hat{x}) $ to input power $(\hat{u})$, is described,
\begin{gather}
    \eta = - \frac{ r_x dx}{ r_u du} = \begin{cases} \tfrac{1}{\eta_b} = 1+\mu \tan \alpha & (\textrm{BWD}),\\ \eta_f = {1-\mu \tan \alpha }  & (\textrm{FWD}).\end{cases} \label{effdef}
\end{gather}
This result aligns with the standard description of the bidirectional efficiency of geared transmissions that is a function of the friction coefficient, contact geometry, and the direction of energy flow \cite{BilateralGear}. 

\subsection{Model Reduction and Efficiency-Null}
First, since coordinates $x$ and $u$ in \eqref{eqn:exDyn} are related by the constraint $g$, the redundant coordinates $\bm s$ are projected onto a minimal coordinate $x$ using the constraint nullspace matrix {$\bM K=[1\quad \sec\alpha]^\top$} such that $\bM{AK}=0$, 
\begin{gather*}
    d\bm s = \bM K dx.
\end{gather*}
Next, the Lagrangian multiplier $\lambda$ is cancelled out from \eqref{eqn:exDyn}. 
Conservative formulations assume that constraints do not dissipate power because the constraint force and its tangent motion are {orthogonal}  \cite{LagrangianManifold}, \cite{TangentBundle}. For dissipative systems, we find an alternative nullity by rearranging \eqref{effdef} and using an efficiency matrix $\bM E_r \!\coloneqq\! \diag{(1, \eta)}$,
\begin{gather}
    r_x dx + \eta r_u du = \bmat{dx\\ du}^{\!\top\!\!} \!\underbrace{\bmat{1 \!&\! 0 \\0 \!&\! \eta}}_{= \bM E_r} \!\bmat{r_x \\ r_u} = dx \bM K^\top \bM E_r \bm r = 0.  \label{dZ}
\end{gather}
We formalize this nullity as \textit{efficiency-null} ($\delta Z \!=\! 0$) which defines the orthogonality between the meshing force and its tangent motion, 
\begin{gather}
    \delta Z \coloneqq  \bM K^\top \bM E_r \bm r = 0. \label{orth}
\end{gather}

\subsection{Asymmetric Dynamics and Mechanical Impedance}

Finally, by multiplying $\bM K^\top \bM E_r $ on both sides of \eqref{eqn:exDyn} and applying $\ddot{\bm s}=\bM K \ddot x $, the meshing force or the Lagrangian multiplier is cancelled out. As a result, we obtain the dynamics of an 1-DoF model:
\begin{gather*}
    \bM K^\top \bM E_r \bm r = \bM K^\top \bM E_r ( \bM H_s \bM K \ddot x - \bm f ) = 0,
\end{gather*}

\begin{align}
    \textrm{FWD:}& \:\: \left(M + \eta_f \frac{m}{\cos^2 \alpha}\vphantom{\frac{1}{\eta_b}}\right)\ddot{x} = f_x-\eta_f\hat{f}_u \label{eqn:exDynFwdFinal},
    \\
    \textrm{BWD:}& \:\:\left( M + \frac{1}{\eta_b} \frac{m}{\cos^2 \alpha}\right)\ddot{x} = f_x - \frac{1}{\eta_b}\hat{f}_u, \label{eqn:exDynBwdFinal}
\end{align}
where $\hat{f}_u$ is the force on $u-$direction, $f_u$, projected onto the $\hat{x}$ coordinate frame, $\hat{f}_u = \tfrac{f_u}{\cos\alpha}$. We utilize the Laplace transform to obtain the mechanical impedance $\bM X(s)$ in the frequency domain $s$. We assume that $\hat{f}_u$ is the only force applied in the FWD case and $f_x$ is the only force exerted in the BWD case.
\begin{IEEEeqnarray}{crCl}
    \textrm{FWD:}&\quad \bM{X}_f (s)=\frac{\hat{\bM f}_u (s)}{\dot {\bM x} (s)} &=& \left( \frac{1}{\eta_f}M + \frac{m}{\cos ^2 \alpha }\right)s, \label{eqn:ForwardImpedance}\\[3pt]
    \textrm{BWD:}&\quad \bM{X}_b (s)=\frac{{\bM f}_x (s)}{\dot{\bM x} (s)}  &=& \left( M + \frac{1}{\eta_b} \frac{m}{\cos ^2 \alpha }\right)s. \label{eqn:BackwardImpedance}
\end{IEEEeqnarray}

\subsection{Discussion}
The results from the wedge-block model highlights the fundamental asymmetry of dissipative dynamics; FWD and BWD dynamics are differently affected by friction $\mu$ and gear ratio $\tfrac{1}{\cos\alpha}$. The {unique properties of the dissipative dynamics of the wedge-block model are summarized as follows}:
\begin{itemize}
    \item \textit{Efficiency}: The backward efficiency $\eta_b$ is always smaller than forward efficiency $\eta_f$. And both are negatively affected by larger gear ratio and friction.
    \item \textit{Non-backdrivability}: There is a limiting case where the system becomes non-backdrivable, $\mu \tan\alpha\!=\!1$ ($\eta_b\!=\!0$).
    \item \textit{Apparent inertia}: \textit{Both} the large gear ratio and the low backward efficiency increases apparent inertia in the BWD case. 
    \item \textit{Impedance}: For both the FWD and BWD cases, the impedance of the system increases with the degradation of mechanical efficiency. 
\end{itemize}
The results demonstrate that the overall BWD dynamics are more negatively affected {by low efficiency}. This occurs because, first, the apparent inertia of the wedge is proportional to the inverse of the backward efficiency. Second, the right hand side of \eqref{eqn:exDynBwdFinal} illustrates that any unmodeled friction in the transmission, such as stiction, is amplified by the gear ratio \textit{and} the inverse of the backward efficiency to resist external forces, which degrades the backdrivability of the mechanism. We note that these properties are also observed in the geared transmissions as reported in \cite{BilateralGear, CycloidvsHarmonic, ImprovingBackdrivability}. {Extending the discussion to the rotor-manipulator model, the negative impact of low transmission efficiency is more likely to affect robots with light-weight links and high gear ratios. }



\section{Generalization of the Dissipative Dynamics}
\label{sec_dissipativedyn}
This section introduces mathematical terms and a graphical tool that are useful to generalize a multi-DoF robotic system with transmissions. First, the angles of rotor, motor, and joint are defined in addition to the mappings between them. Next, we conceptualize transmissions by formulating kinematic constraints that determine the mechanical connectivity between rigid bodies. Finally, we trace the frictional losses in the transmissions and discuss the topological propagation of power.

\subsection{Kinematic Topology and Constraints}

This section provides the mathematical definitions for two core concepts: the speed reduction and the actuation topology. Speed reduction is often realized by employing pairs of meshed gears or belt-driven pulleys with different diameters. The actuation topology describes the topology of the mechanism that distributes power from the motors to the joints. The actuation topology, or the kinematic structure \cite{SerialParallelComparison}, is used to classify robots as, for example, serial or parallel mechanisms. We define these two concepts as transformations between coordinate angles. For that, three sequential states (angular positions) are introduced:
\begin{itemize}
    \item \textit{Rotor angle} $\bm \phi$ is the angular position of the motor rotor before a gearbox or a reduction mechanism;
    \item \textit{{Motor} angle} $ \bm \psi$ is the angular position of the output of the actuator after a reduction mechanism; and
    \item \textit{Joint angle} $ \bm q $ is the angle between the structural links of the robotic system. 
\end{itemize}

The mappings between the states are defined as:
\begin{itemize}
    \item \textit{Reduction} $\bM R$ is the generalization of the speed reduction, which maps the displacement of the rotor angle $\bm \phi$ to the motor angle $ \bm \psi$; and 
    \item \textit{Actuation topology} $\bM D$ represents the kinematic topology of a transmission, which maps the motor angle $ \bm \psi$ to the joint angle $ \bm q $.
\end{itemize}

\begin{figure}[t]
  \centering
  \includegraphics[width =\linewidth]{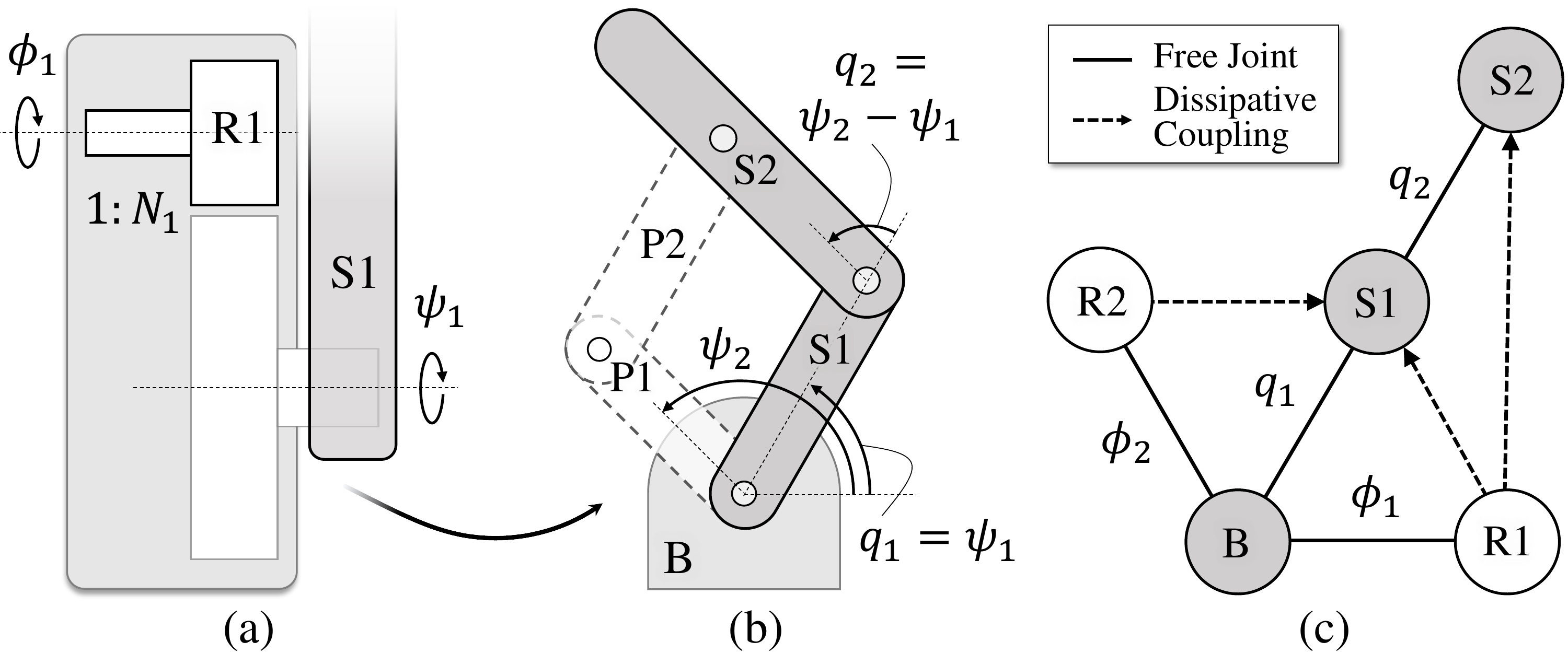}
  \caption{The coordinate angles in a 2-DoF robotic arm are visualized. 
(a) A single stage gearbox in an actuator (b) A 2-DoF robotic arm using parallelogram (c) The kinematic tree outlines the topology of a transmission and the kinematic constraints associated with power losses.}
  \label{Fig:RigidBody}
\end{figure}

Fig. \ref{Fig:RigidBody}(a) and (b) illustrates an {actuation topology} of a typical 2-DoF parallelogram mechanism in which the speed reduction is represented by a single-stage gearbox. Similar mechanisms have been used in the legged robots ATRIAS \cite{ATRIAS_Design} and Minitaur \cite{Minitaur_DesignPrinciples}. The two motors are fixed to the base and drive the links $\textrm{S1}$ and $\textrm{P1}$ of the parallelogram mechanism. As a consequence, the input power propagates to the end-effector via links $\textrm{P2}$ and $\textrm{S2}$.
Each motor has a gearbox that reduces the rotation of the rotor to motor angle $\psi_i\!=\!\tfrac{\phi_i}{N_i}$ for $i \!\in\! \{1,2\}$, where ${N_i}$ {are gear ratios}. The first joint is driven by the first actuator ($q_1\!=\!\psi_1$), while the second joint angle is controlled by both motors ($q_2\!=\!\psi_2\!-\!\psi_1$). Following our previous definitions, the reduction and actuation topology of this manipulator are given by:
\begin{IEEEeqnarray*}{RCL} 
\bM R_{\rm par} = 
\left[\begin{IEEEeqnarraybox*}[][c]{,c/c,}
\tfrac{1}{N_1} & 0 \\
0 & \tfrac{1}{N_2}
\end{IEEEeqnarraybox*}\right],\quad \quad
\bM D_{\rm par} = 
\left[\begin{IEEEeqnarraybox*}[][c]{,c/c,}
1 & 0 \\
-1 & 1
\end{IEEEeqnarraybox*}\right].
\end{IEEEeqnarray*}
If the actuation topology represented a serial mechanism in which the motors are directly mounted on the joints, the matrix $\bM D_\textrm{ser}$ would be $2\times2$ identity matrix, while the reduction matrix would remain unchanged, {$\bM R_\textrm{ser}\!=\!\bM R_\textrm{par}$}. 
Next, the kinematic constraints in actuators and transmissions, are represented with the reduction and actuation topology matrices:
\begin{IEEEeqnarray*}{rCl}
    \bm g(\bm q, \,\bm \phi) = \bm q - \bM{D R} \bm \phi. 
\end{IEEEeqnarray*}

Finally, we include the task-space coordinate of the end-effector $\bm x$ and its Jacobian $\bM J$ to obtain the sequential transformations of generalized coordinates,
\begin{IEEEeqnarray}{CCCCCCC}
    d \bm \phi &\xrightarrow{\:\:\bM R\:\:} &d \bm \psi &\xrightarrow{\:\:\bM D\:\:}& d \bm q &\xrightarrow{\:\:\bM J\:\:}& d \bm x. \label{DualFlowMotion}
\end{IEEEeqnarray}
To simplify the discussion, it is assumed that the matrices $\bM R$ and $\bM D$ are dimensionless and invertible. Readers can refer to \cite{SerialParallelComparison, GosselinSingularityAnalysis, MultiLoopKinematics} for the analysis of nonlinear constraints. 

\subsection{Rigid Body Systems and Kinematic Tree}
We employ the kinematic tree of a rigid body system \cite{FeatherstoneBook} to trace the energy losses associated with the kinematic constraints in a transmission mechanism. Mechanical connections between rigid bodies are categorized into two types: (i) free joints, which define the parent-child relations and (ii) dissipative couplings, which represent a kinematic constraint that bears power loss. For example, Fig. \ref{Fig:RigidBody}(c) is the kinematic tree of a parallelogram mechanism, showing energy transfer from rotors $\textrm{(R1, R2)}$ to structural bodies $\textrm{(S1, S2)}$. The dissipative couplings are represented by dashed arrows to clearly illustrate that the power losses depend on the direction of power delivery.
In order to avoid kinematic loops that would complicate the analysis of the dynamics, we assume that some components, such as $\textrm{P1}$ and $\textrm{P2}$, in a transmission are massless. 


\section{{The Dynamics of Dissipative Rigid-Body Systems}}
{The dissipative dynamics of a general robotic system is obtained similarly to the solution for the simple example using the definitions from section \ref{sec_dissipativedyn}. The goals are to (i) create a kinematic tree of rigid-bodies with redundant states, (ii) define kinematic couplings between rigid-bodies written as constraints, (iii) collect meshing forces given by dissipative forces and constraint forces, and (iv) reduce the model by projecting its dynamics onto the tangent space to eliminate the meshing forces.}

The general model of a legged robot is comprised of a $b$-DoF floating-base (the torso) and a serial chain of $m$-links (the limbs) \cite{HumanoidDynamics}. The limbs are driven by $m$-rotors through mechanical transmissions as defined by a reduction matrix $\bM R_m$ and an actuator topology matrix $\bM D_m$. 
We define $\bm q_b\in \R^{b}$ as the $b$-DoF coordinates for the free floating-base, and $\bm \phi_m, \bm q_m\in \R^{m}$ as the rotor angles and joint angles. Next, the constraint in the transmission is denoted as
\begin{equation}
\bm g(\bm q_m, \bm \phi_m ) = \bm q_m -\bM D_m \bM R_m \bm \phi_m.    \label{eqn:genConstr}
\end{equation}
The generalized coordinates of the redundant model and the reduced-order model are $\bm s = [\bm q_b ^\top, \bm q_m ^\top, \bm \phi_m^\top ]^\top$ and $\bm q = [\bm q_b ^\top, \bm q_m ^\top ]^\top$. The constraint Jacobian $\bM A$ and its nullspace matrix $\bM K$ such that $d\bm s \!=\! \bM K d\bm q$ and that $\bM A\bM K \!=\! \bm 0 $ are written
\begin{align*}
    \bM A = \mpdif{\bm{g}}{\bm s} = \bmat{\; \bm 0_{m\times b} & \bm1_{m}& -\bM{D}_m \bM{R}_m \;\;} \in \R^{m\times(b+2m)},
    \\
    \bM K = \bmat{\;\;\bm 1_b & \bm 0 \\ \;\;\bm 0 & \bm 1_m \\ \;\; \bM 0 & (\bM{D}_m \bM{R}_m)^{-1}} \in \R^{(b+2m)\times (b+m)},
\end{align*}
{where $\bm1_{n}$ is the $n\times n$ identity for any $n\in \mathbb{N}$.} The equation of motion of the redundant system is given as
\begin{IEEEeqnarray}{rCl}
    \bM H_s \ddot{\bm s} +\bm c_s -\bM A^\top \bm \lambda_m = \bm \tau_s + \bm \tau_d, \label{redundantSysEq}
\end{IEEEeqnarray}
where $\bM H_s$ and $\bm c_s$  are inertia matrix and bias force, $\bm \lambda_m \!\in\! \R^{m}$ are the Lagrange multipliers, $\bm \tau_s$ are the generalized forces, and $\bm \tau_d \!= \![0, \! \cdots \!, 0, \tau_{d,1}, \!\cdots \!, \tau_{d,m}]^\top \! \in \! \R^{b+2m}$ are generalized dissipative forces in the transmission. 
The forces due to the constraints are combined as meshing forces 
\begin{gather}
    \bm r = \bM A^\top \bm \lambda_m + \bm \tau_d = \bM H_s \ddot{\bm s} + \bm c_s -\bm \tau_s. 
\end{gather}
We assume that for all $m$-kinematic couplings of power efficiencies $\eta_1, \cdots, \eta_m$, their associated efficiency-nulls $\delta Z_1, \cdots, \delta Z_m$ are all equal to zero. By stacking efficiency-nulls, we obtain an orthogonality between the meshing force and its tangent motion at system-level,
\begin{gather}
    \bM K^\top \bM E_s \bm r =\bM 0, \label{sysdZ}
\end{gather}
where $\bM E_s \!\coloneqq \!\diag{(1, \!\cdots\!, 1, \eta_1, \!\cdots\!, \eta_m)}\!\in\! \R^{(b+2m)\times(b+2m)}$. 

The equation of the motion of the redundant system \eqref{redundantSysEq} is reduced to $(b+m)$-dimensional system by left multiplying $\bM{K}^\top \bM{E}_s$ to the meshing force $\bm r$, and projecting $\ddot {\bm s}$ onto the reduced coordinates $\ddot {\bm q}$ with $\ddot{\bm s} = \bM K \ddot {\bm q}$. This procedure yields a \textit{dissipative equation of motion} of a robotic system,
\begin{gather}
    \bM K^\top \bM E_s \bM H_s \bM K\ddot{\bm q}  + \bM{K}^\top \bM E_s \bm c_s = \bM J^\top \bm f_{\textrm{ext}} + \bmat{\bm 0_{b\times m} \\ \bM B_m \bM E_m}\bm\tau_\phi,  \label{eqn:dissipativeDyn}
\end{gather}
where $\bM J \!\in\! \R^{n \times(b+m) }$ is the contact Jacobian of an external force at end-effector $\bm f_{\textrm{ext}}\!\!\in\! \R^n$ , $\bM B_m \!\coloneqq\! \pmat{\bM D_m \bM R_m}^{\!-\!\top}$ is a distribution matrix, $\bM E_m \!\coloneqq\! \diag{(\eta_1, \!\cdots\!, \eta_m)}$ is the efficiency matrix, and $\bm \tau_{\phi}\!\in \! \R^m$ are torques applied to the rotors. We assume that there are no forces applied to the floating-base and {that each end-effector contacts the environment at a single point.} The result takes the form of the conventional manipulator equations of motion \cite{ModernRobotics}, with the transmission efficiencies embedded into the inertia and Coriolis matrices, and generalized forces term. 

\subsection{Application of the formulation}
In contrast with the conservative formulation, the dissipative equation of motion makes evident the concept of energy flow in the system. There are different expressions if the actuators are driving the robot with torques $\bm \tau_\phi$ (FWD) or if the limbs are being accelerated by external forces $\bm f_{\textrm{ext}}$ (BWD). This concept of asymmetric transmission dynamics leads to separate design criteria if the robot actuators are expected to perform positive or negative work, and hence, conventional metrics can be re-derived \cite{HandbookofRoboticsDesignCriteria, DynamicCapability}. We illustrate this characteristic by analyzing the design of a 2-DoF leg and compute its generalized inertia ellipsoid  \cite{AsadaInertiaEllipsoid}, its force capability or tip force bounds \cite{ModernRobotics}, and its ability to mitigate shock loads due to impacts between the foot and the ground. The latter is defined by the Impact Mitigation Factor (IMF) \cite{ImpactMitigation}, which ranges from zero to one and quantifies the inertial backdrivability of the mechanism. 

\subsubsection{Generalized Inertia Tensor}
The generalized inertia tensor (GIT) or the generalized inertia ellipsoid describes the inertia felt at the end-effector frame of a robot. We propose that \textit{the generalized inertia tensor of a dissipative system depends on the efficiency of transmissions and the direction of energy flow}. Due to the dependency of the transmission dynamics on the direction of energy flow, the end-effector inertia perceived by the motors will be different than the inertia perceived by an external force back-driving the robot. The calculation of \textit{Backward-GIT} is identical to the conventional GIT, since the external force is not distorted by transmission efficiency. However, the propagation of the actuation torque is different in the case of \textit{Forward-GIT}. To accelerate the end-effector, the actuators exert a virtual task-space force $\hat{\bm f}_{\textrm{task}}$ given by 
\begin{gather}
    \bm{\tau_\phi} = \bM B_m^{-1} \bM J ^\top {\bm {\hat{f}_{\textrm{task}}}}. \label{VirtualForce}
\end{gather}
For the above equation to hold in the forward-driving cases, it is assumed that the system is fully-actuated or fixed-based ($b=0$). 
Equation \eqref{eqn:dissipativeDyn} provides and \eqref{VirtualForce} the Forward-GIT. 
\begin{IEEEeqnarray*}{cCl}
    \textrm{Backward-GIT} &\coloneqq&  \big( {\bM{ J} {\bM H}^{-1}  \bM{ J}^{\top} } \big)^{-1},
    \\
    \textrm{Forward-GIT} &\coloneqq & \big({ \bM{ J} {\bM H}^{-1}  \bM B_m \bM E_m \bM B_m ^{-1}{\bM{ J}}^{\top} }\big )^{-1},
\end{IEEEeqnarray*}
where $\bM H \!=\! \bM K ^\top \bM E_s \bM H_s \bM K$. 
Analogous to the forward and backward impedance described in equations \eqref{eqn:ForwardImpedance} and \eqref{eqn:BackwardImpedance}, the Forward- and Backward-GIT are larger in respect to the GIT due to the lower efficiency of transmission.

\subsubsection{Task-Space Force Capability}
The task-space force capability (FC) estimates the maximum contact force that the robot can produce at the end-effector. Conventionally, this concept shows that this force is limited by the maximum torque that the actuators can generate. However, \textit{we propose that FC also depends on the mechanical efficiency of the robot's transmission.} To better understand this concept, assume that a legged robot semi-statically interacts with the ground to support it's own body weight. In the forward-driving case, the leg motors must overcome the forces due to the robot mass plus the dissipative forces in the transmission. Hence, the lower the efficiency of the transmission, the higher the motor input effort must be to lift the robot. However, in the BWD case, the gravitational torques due to the robot weight must drive the motors through the transmission. Interestingly, in this scenario, the lower efficiency (due to high friction) \textit{helps} the robot to passively support its own body. Thus, legged robots or industrial manipulators which employ highly geared and low efficiency actuators do not collapse under their own body weight when powered off. This phenomena is captured by \textit{Asymmetric-FC} that combines Forward-FC and Backward-FC . The Asymmetric-FC estimates the robot's ability to quasi-statically support or resist external forces, 
\begin{gather}
\textrm{Asymmetric-FC} =      {\bM J_m^{-\top}} \bM B_m \bM E_m \convhull({\bm \tau}_{{\phi}}),
\end{gather}
where we assume that the torque on rotors $\bm \tau_\phi$ is bounded and its feasible region is represented by a convex-hull. The Asymmetric-FC only employs the limb's contact Jacobian $\bM J_m \in \R ^{3\times m}$, and not that of the whole system. If a  robot has redundant actuators or is in a singular pose, the Asymmetric-FC can be obtained by linear programming \cite{ForceCapabilityPolytope}.

\subsection{Effect of dissipative dynamics}
We present two analysis of the design of a 2-DoF  planar leg composed of a serial actuation mechanism. First, we investigate the effect of transmission efficiencies on the task-space (foot) inertia matrix using the Forward- and Backward-GIT, and the force capability using the Asymmetric-FC. Next, the Asymmetric-FC and the IMF in the vertical $\hat z$ direction is computed to analyze the dynamic response of the leg to external contact forces. The Asymmetric-FC is normalized by the conventional FC to measure the relative difference due to the internal friction in the transmissions. 

\begin{figure}[t]
  \centering
  \includegraphics[width =1\linewidth]{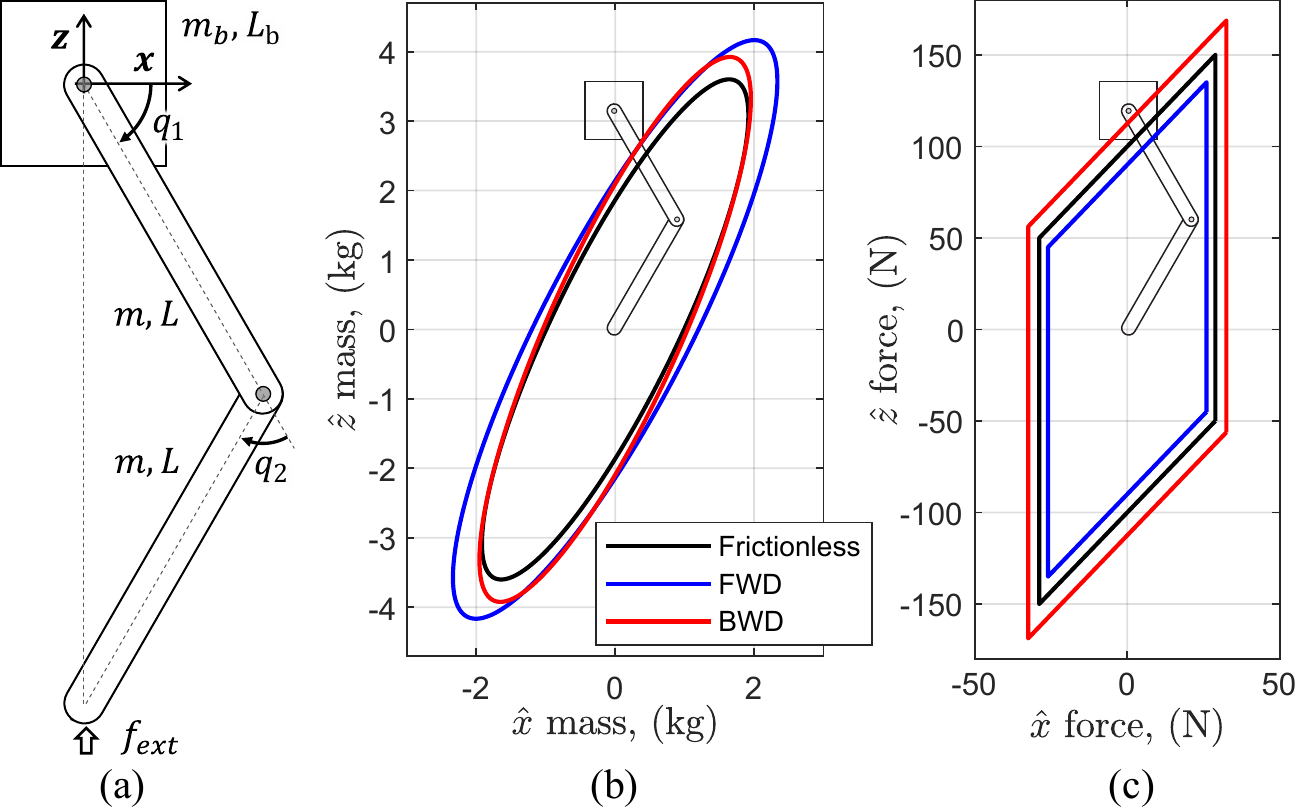}
  \caption{(a) The 2-DoF floating base robot for the design study (b) The forward and backward generalized inertia ellipsoids of the 2-DoF Leg (c) The asymmetric force capabilities of the 2-DoF Leg}
  \label{Fig:GITandFC}
\end{figure}

\begin{figure}[t]
  \centering
  \includegraphics[width =1\linewidth]{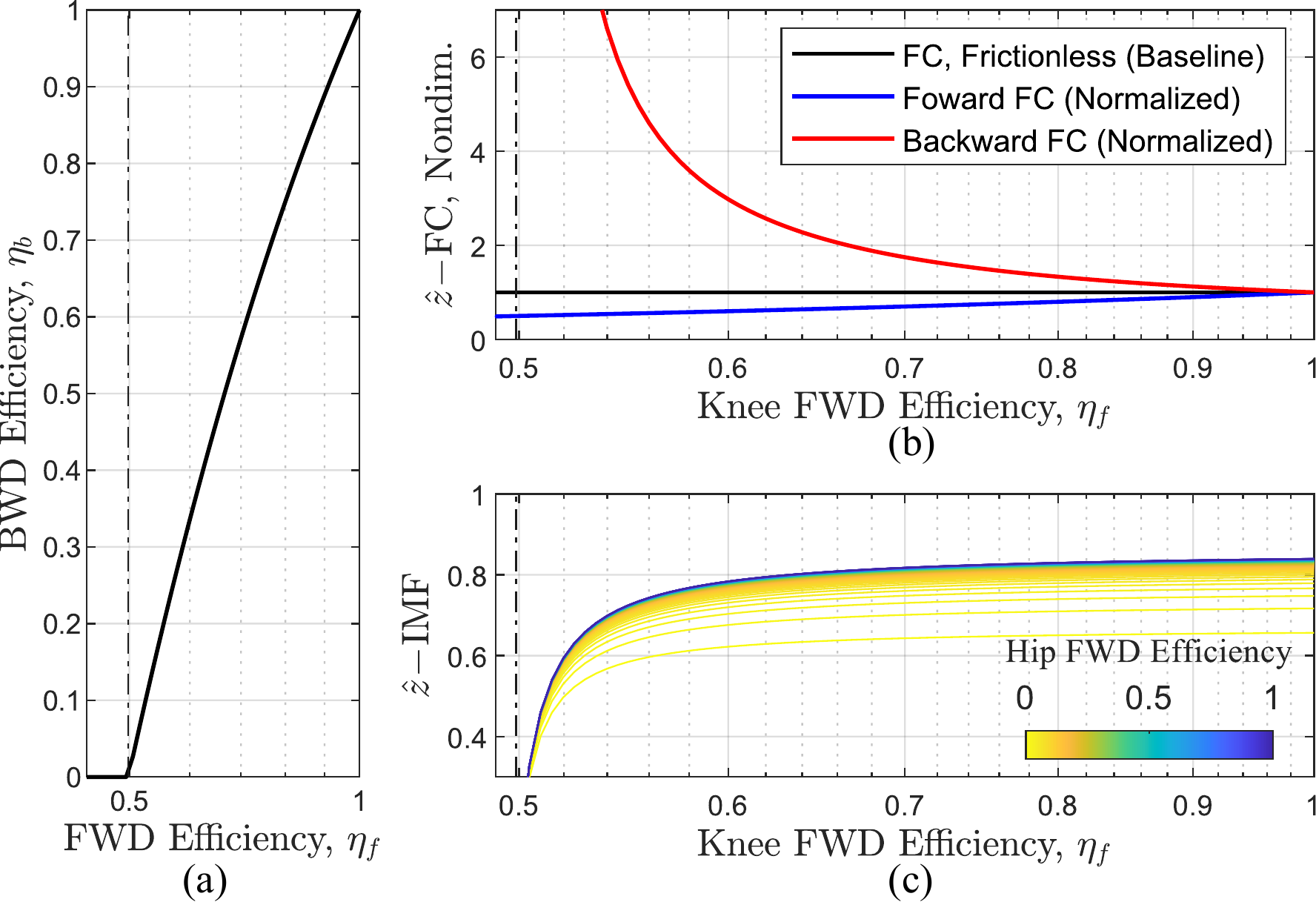}
  \caption{ 
  (a) The relation between forward and backward efficiencies. The backward efficiency converges to zero at $\eta_f=0.499$. (b) The $\hat{z}$-directional forward and backward force capabilities normalized by force capability. (c) The $\hat{z}$-directional IMF of 2-DoF Leg}
  \label{Fig:IMFvsForceCapability}
\end{figure}

The leg, shown in {Fig. \ref{Fig:GITandFC}(a)}, is equipped with two identical motors that can exert up to $20 \textrm{ N}\!\cdot \!\textrm{m}$ after identical speed reduction of $20:1$. Motors are attached to the hip ($ q_1$) and the knee ($ q_2$) joint. The thigh and the shin have the identical mass $m\!=\!2\textrm{ kg}$ and length $L\!=\!0.4\textrm{ m}$. Their center of mass are located at the links' midpoint. The base is modeled as a uniform planar square of dimensions $L_b\!=\! 0.5\textrm{ m}$ and mass $m_b\!=\!15\textrm{ kg}$. The base and the motors are initially at rest, with joint angles, {$q_1\!=\!q_2\!=\frac{\pi}{3}$}. The reflected inertia of a rotor is set identical to the inertia of a connected link at its center of mass. We assume a typical forward efficiency of single-stage gearboxes of $0.8$ and $0.7$ for the hip and knee transmission. The backward efficiency is a function of forward efficiency $\eta_f$ as displayed in Fig. {\ref{Fig:IMFvsForceCapability}}(a) and calculated by the equation (38) of \cite{BilateralGear}. 
We note that the backward efficiency is \textit{always} smaller than the forward efficiency, and converges to zero when $\eta_f \xrightarrow{} 0.499$. 

We observe the behavior of the asymmetric transmission dynamics in Fig. \ref{Fig:GITandFC}. First, both ellipsoids of Forward- and Backward-GIT in Fig. \ref{Fig:GITandFC}(b) are always larger than that of GIT. 
Intuitively, driving forces need to inject more {energy} into the system to compensate for the frictional losses in the transmissions, which renders larger inertia. 
In contrast, the Forward- and the Backward-FC in Fig. \ref{Fig:GITandFC}(c) show opposite tendencies. As the transmission efficiency decreases, the Forward-FC linearly decreases while Backward-FC diverges to infinity when $\eta_f\xrightarrow{}0.499$. Consequently, the leg can quasi-statically withstand more substantial load. 

Finally, Fig. \ref{Fig:IMFvsForceCapability} shows the FC, Forward- and Backward-FC, and IMF of the foot in the vertical $\hat z$ direction. The result conveys the \textit{trade-off} between the Backward-FC and the IMF. As the actuators' efficiency decrease, the $\hat z-$Backward-FC in Fig. \ref{Fig:IMFvsForceCapability}(b) increases, while $\hat z-$IMF in Fig. \ref{Fig:IMFvsForceCapability}(c) decreases. In other words, large mechanical losses in the transmissions allow the robot to sustain more substantial static forces, but also degrade the machine's ability to mitigate shock loads from impacts with the ground. 

\section{Conclusion}
This paper investigates how the dissipative forces in actuators and transmissions propagate to the dynamics of a whole robot. We present a framework that uses the mechanical efficiency to augment the inertia, Coriolis, and generalized force terms in the equation of motion. 
{We show how the individual efficiency of transmission influences not only the inertia felt at the end-effector, but also its capability of applying forces to and resisting disturbances from the environment. We expect that roboticists will use this formulation to add a tunable variable, the mechanical efficiency, to optimize the design of multi-body systems. }
For instance, designers may exploit the beneficial effects of low efficiency to gain more static load-bearing capability, or to minimize the negative effects of friction for dynamic tasks with impacts. 






\bibliographystyle{ieeetr}
\bibliography{bib_ywsim.bib}

\begin{thebibliography}{10}

\bibitem{ImpactMitigation}
P.~M. {Wensing}, A.~{Wang}, S.~{Seok}, D.~{Otten}, J.~{Lang}, and S.~{Kim},
  ``Proprioceptive actuator design in the mit cheetah: Impact mitigation and
  high-bandwidth physical interaction for dynamic legged robots,'' {\em IEEE
  Transactions on Robotics}, vol.~33, no.~3, pp.~509--522, 2017.

\bibitem{MiniCheetah}
B.~{Katz}, J.~D. {Carlo}, and S.~{Kim}, ``Mini cheetah: A platform for pushing
  the limits of dynamic quadruped control,'' in {\em 2019 International
  Conference on Robotics and Automation (ICRA)}, pp.~6295--6301, 2019.

\bibitem{Ramos_LilHERMES}
J.~{Ramos}, B.~{Katz}, M.~Y.~M. {Chuah}, and S.~{Kim}, ``Facilitating
  model-based control through software-hardware co-design,'' in {\em 2018 IEEE
  International Conference on Robotics and Automation (ICRA)}, pp.~566--572,
  2018.

\bibitem{TELLO_Vid_Leg}
Y. Sim, TELLO's Leg Mechanism Test. (Aug. 23, 2020). (Accessed: Jan. 7. 2021.
  [Online Video]. Available: \url{https://youtu.be/R0_2LmV3WQo}.

\bibitem{ChoiceMotorReducer}
H.~Giberti, S.~Cinquemani, and G.~Legnani, ``Effects of transmission mechanical
  characteristics on the choice of a motor-reducer,'' {\em Mechatronics},
  vol.~20, no.~5, pp.~604 -- 610, 2010.

\bibitem{Wang_DirectionalEfficiency}
A.~{Wang} and S.~{Kim}, ``Directional efficiency in geared transmissions:
  Characterization of backdrivability towards improved proprioceptive
  control,'' in {\em 2015 IEEE International Conference on Robotics and
  Automation (ICRA)}, pp.~1055--1062, 2015.

\bibitem{SinghShockPropagation}
B.~R.~P. {Singh} and R.~{Featherstone}, ``Mechanical shock propagation
  reduction in robot legs,'' {\em IEEE Robotics and Automation Letters},
  vol.~5, no.~2, pp.~1183--1190, 2020.

\bibitem{LIMS}
Y.~{Kim}, ``Anthropomorphic low-inertia high-stiffness manipulator for
  high-speed safe interaction,'' {\em IEEE Transactions on Robotics}, vol.~33,
  no.~6, pp.~1358--1374, 2017.

\bibitem{YadaGearReview}
T.~Yada, ``Review of gear efficiency equation and force treatment,'' {\em JSME
  international journal. Ser. C, Dynamics, control, robotics, design and
  manufacturing}, vol.~40, no.~1, pp.~1--8, 1997.

\bibitem{AsadaInertiaEllipsoid}
H.~{Asada}, ``Dynamic analysis and design of robot manipulators using inertia
  ellipsoids,'' in {\em Proceedings. 1984 IEEE International Conference on
  Robotics and Automation}, vol.~1, pp.~94--102, 1984.

\bibitem{BilateralGear}
H.~{Matsuki}, K.~{Nagano}, and Y.~{Fujimoto}, ``Bilateral drive gear—a highly
  backdrivable reduction gearbox for robotic actuators,'' {\em IEEE/ASME
  Transactions on Mechatronics}, vol.~24, no.~6, pp.~2661--2673, 2019.

\bibitem{LagrangianManifold}
V.~I. Arnol'd, {\em Mathematical methods of classical mechanics}, vol.~60.
\newblock Springer Science \& Business Media, 2013.

\bibitem{TangentBundle}
M.~Crampin, ``Tangent bundle geometry lagrangian dynamics,'' {\em Journal of
  Physics A: Mathematical and General}, vol.~16, pp.~3755--3772, nov 1983.

\bibitem{CycloidvsHarmonic}
J.~W. {Sensinger} and J.~H. {Lipsey}, ``Cycloid vs. harmonic drives for use in
  high ratio, single stage robotic transmissions,'' in {\em 2012 IEEE
  International Conference on Robotics and Automation}, pp.~4130--4135, 2012.

\bibitem{ImprovingBackdrivability}
T.~Nef and P.~Lum, ``Improving backdrivability in geared rehabilitation
  robots,'' {\em Medical \& biological engineering \& computing}, vol.~47,
  no.~4, pp.~441--447, 2009.

\bibitem{SerialParallelComparison}
Z.~Pandilov and V.~Dukovski, ``Comparison of the characteristics between serial
  and parallel robots.,'' {\em Acta Technica Corvininesis-Bulletin of
  Engineering}, vol.~7, no.~1, 2014.

\bibitem{ATRIAS_Design}
C.~Hubicki, J.~Grimes, M.~Jones, D.~Renjewski, A.~Spröwitz, A.~Abate, and
  J.~Hurst, ``Atrias: Design and validation of a tether-free 3d-capable
  spring-mass bipedal robot,'' {\em The International Journal of Robotics
  Research}, vol.~35, no.~12, pp.~1497--1521, 2016.

\bibitem{Minitaur_DesignPrinciples}
G.~{Kenneally}, A.~{De}, and D.~E. {Koditschek}, ``Design principles for a
  family of direct-drive legged robots,'' {\em IEEE Robotics and Automation
  Letters}, vol.~1, no.~2, pp.~900--907, 2016.

\bibitem{GosselinSingularityAnalysis}
C.~{Gosselin} and J.~{Angeles}, ``Singularity analysis of closed-loop kinematic
  chains,'' {\em IEEE Transactions on Robotics and Automation}, vol.~6, no.~3,
  pp.~281--290, 1990.

\bibitem{MultiLoopKinematics}
K.~Tchoń, ``Differential topology of the inverse kinematic problem for
  redundant robot manipulators,'' {\em The International Journal of Robotics
  Research}, vol.~10, no.~5, pp.~492--504, 1991.

\bibitem{FeatherstoneBook}
R.~Featherstone, {\em Rigid body dynamics algorithms}.
\newblock Springer, 2014.

\bibitem{HumanoidDynamics}
T.~Sugihara and M.~Morisawa, ``A survey: dynamics of humanoid robots,'' {\em
  Advanced Robotics}, vol.~34, no.~21-22, pp.~1338--1352, 2020.

\bibitem{ModernRobotics}
K.~M. Lynch and F.~C. Park, {\em Modern Robotics}.
\newblock Cambridge University Press, 2017.

\bibitem{HandbookofRoboticsDesignCriteria}
O.~Khatib and B.~Siciliano, {\em Springer handbook of robotics}.
\newblock Springer International Publishing, 2016.

\bibitem{DynamicCapability}
A.~{Bowling} and O.~{Khatib}, ``The dynamic capability equations: a new tool
  for analyzing robotic manipulator performance,'' {\em IEEE Transactions on
  Robotics}, vol.~21, no.~1, pp.~115--123, 2005.

\bibitem{ForceCapabilityPolytope}
P.~{Chiacchio}, Y.~{Bouffard-Vercelli}, and F.~{Pierrot}, ``Evaluation of force
  capabilities for redundant manipulators,'' in {\em Proceedings of IEEE
  International Conference on Robotics and Automation}, vol.~4, pp.~3520--3525
  vol.4, 1996.

\end{thebibliography}

\end{document}